%% file: GRUD-ODE.tex
\documentclass[letterpaper]{article} 
\usepackage{aaai21}  
\usepackage{times}  
\usepackage{helvet} 
\usepackage{courier}  
\usepackage[hyphens]{url}  
\usepackage{graphicx} 
\usepackage[numbers, sort&compress, square]{natbib}  
\usepackage{caption} 
\usepackage[backref]{hyperref} 
\hypersetup{hidelinks}
\usepackage{amsmath,amssymb,amsfonts,bm}
\usepackage{algorithmic}
\usepackage{xcolor}
\usepackage[linesnumbered,ruled,vlined]{algorithm2e}
\usepackage{mathtools}
\usepackage{multirow}
\usepackage{booktabs}
\usepackage{stackengine}
\usepackage{subfig}
\usepackage{graphicx}
\usepackage{caption}

\usepackage{longtable}

\usepackage{array}
\usepackage{booktabs}

\usepackage{multirow, boldline}

\usepackage{makecell} 

\usepackage{hyperref}

\usepackage{arydshln} 

\usepackage{booktabs}

\usepackage{multirow}
\usepackage{listings}

\begin{document}
%
\input{defs}
\title{GRU-TV: Time- and velocity-aware GRU for patient representation on multivariate clinical time-series data}
\author{
    Ningtao Liu\textsuperscript{\rm 1, \rm 2}, Ruoxi Gao\textsuperscript{\rm 3}, and Shuiping Gou\textsuperscript{\rm 1,*}
    \\
}
\affiliations{
    \textsuperscript{\rm 1}Key Laboratory of Intelligent Perception and Image Understanding of Ministry of Education, School of Artificial Intelligence, Xidian University, Xi’an 710071, China\\
    nt\_liu@stu.xidian.edu.cn\\
    \textsuperscript{\rm 2}Robarts Research Institute, Western University, London, ON, Canada, N6A 5B7\\
    nliu258@uwo.ca\\
    \textsuperscript{\rm 3}Electrical Engineering and Computer Science  Department, University of Michigan, Ann Arbor, MI, 48109 
    gruoxi@umich.edu\\
    \textsuperscript{\rm *}Corresponding author\\
    \textsuperscript{\rm **}These authors contributed equally: Ningtao Liu, Ruoxi Gao\\

}

\maketitle
\begin{abstract}
Electronic health  records (EHRs) are usually highly dimensional, heterogeneous, and multimodal.
Besides, the random recording of clinical variables 
results in high missing rates and uneven time intervals between adjacent records in the multivariate clinical time-series data extracted from EHRs.
Current works using clinical time-series data for patient representation regard the patients' physiological status
as a discrete process described by sporadically collected records. 
However, changes in the patient's physiological condition are continuous and dynamic processes. 
The perception of time and velocity of change is crucial for patient representation learning.

In this study, we propose a time- and velocity-aware gated recurrent unit model (GRU-TV)
for patient representation learning of clinical multivariate time-series data in a time-continuous manner.
The neural ordinary differential equations (ODEs) and velocity perception mechanism are applied to perceive
the time interval between adjacent records and changing rate of the patient's physiological status, respectively.
Our experiments on two real clinical EHR datasets (PhysioNet2012, MIMIC-III) establish that GRU-TV is a robust model
on computer-aided diagnosis (CAD) tasks, especially on sequences with high-variance time intervals.
\end{abstract}

\section{INTRODUCTION}
In recent years, a large number of health care-relevant records ~\cite{johnson2016mimic} are used for computational analysis.
According to the Healthcare Cost and Utilization Project (HCUP) of the National Inpatient Sample, 
in 2018, more than 7 million patients in the United States were hospitalized, and over 10 million medical records are generated
\footnote{https://www.hcup-us.ahrq.gov/db/nation/nis/nissummstats.jsp}.
Clinician decisions on treatment and diagnosis of diseases highly rely on these medical records, 
such as lab tests, previous procedures, medications, and diagnoses. 
A large amount of clinical data provide an opportunity for computer-assisted diagnosis (CAD) applications, specifically those with machine learning methods.

Recently, there are researchers focused on clinical diagnosis tasks using patient representation learning, 
such as patient subtyping~\cite{zhang2018time,baytas2017patient,che2015deep}, 
mortality prediction~\cite{xu2018raim, song2018attend, kemp2019improved, yin2019learning}, 
length-of-stay prediction~\cite{si2019deep, xu2018raim, song2018attend}, 
24-hour decompensation ~\cite{xu2018raim, song2018attend, suresh2018learning},
and multi-task learning~\cite{harutyunyan2019multitask}.

RNN and its variant models such as long short term memory (LSTM), 
and gated recurrent unit (GRU) are widely used for patient representation learning with time-series data in these previous studies.
However, multivariate sequences extracted from electronic health records (EHRs) are usually high-dimensional, heterogeneous, and multimodal~\cite{Si2021},
since the data points in EHRs are typically collected from different resources, including medical sensors and medical laboratories. 
In addition, the missing rate of variable values is high in these sequences.
which results in an irregular missing pattern.
Multiple sources and a high missing rate lead to uneven time intervals between records in the sequence.
These properties result in basic RNNs that do not work well for patient representation learning
because the basic assumptions of RNNs are that the time interval between records in the input sequence is uniform
and the length of the records is consistent.

Improved RNN models~\cite{baytas2017patient, wang2018predictive} adapt to clinical time-series data by integrating time intervals into the model.
These methods require nonlinear mapping of time intervals and represent the patient's physiological state in a discrete manner.
The RNNs with decay term present in~\cite{che2018recurrent, cao2018brits} decay the hidden state in forward propagation by the time decay terms, 
which was proven not to improve performance over standard RNN in~\cite{mozer2017discrete}. 
In addition to improving the basic RNN units, 
some methods applied ordinary differential equation (ODE) to RNN for processing non-uniformly sampled time-series data~\cite{de2019gru, rubanova2019latent}
after the neural ODE was proposed in~\cite{NEURIPS2018_69386f6b}. 



In clinical practice,
instantaneous vital sign rate related to physiological dynamics is an important indicator of the human health condition~\cite{zhao2017noncontact}.
Thus, the perception of changes in a patient's physical condition is also critical for patient representation learning.
Current RNN models, whether representing a patient's physiological status in a continuous or discrete manner, 
lack the ability to perceive the changes in patient's physiological status.

In this work, we propose a Time- and Velocity-aware GRU (GRU-TV) for patient representation based on GRU.
In the proposed GRU-TV, the update of the hidden state and the forward propagation of the RNN along time are parallel, which
interacts through the ODE of the hidden state and the time interval.
Inspired by neural ODE, the forward propagation of hidden states between records is 
improved to perceive the time interval.
Besides, the velocity perceptron is achieved by expanding the instantaneous rate of the hidden state into the gate.


\section{RELATED WORK}
\label{sec:RELATED WORK}
Related works using deep learning models for patient representation learning, such as length of stay prediction~\cite{rajkomar2018scalable}, in-hospital mortality prediction~\cite{lei2018effective},
patient phenotyping~\cite{baytas2017patient}, and international classification of diseases (ICD) code classification~\cite{purushotham2018benchmarking} follow a similar paradigm. 
In this paradigm, the deep learning models are used for representing a patient's physiological condition and are followed a task-specific network for generating specific predictions or classification results.
Therefore, the design of a deep learning model for patient representation based on EHR is critical for performance.
In this section, we review related work from two perspectives, patient representation learning and time interval perception.

\textbf{Patient representation learning}.
In terms of methods, deep learning models are increasingly used in patient representation learning tasks.
These models are generally generic, i.e., task-independent, focusing on the representation of the patient's physiological 
condition from clinical time-series data extracted from the EHR.
Many studies have applied different methods on the basic RNN model to adapt to the sparsity, 
high dimensionality and inhomogeneity of clinical time series data.
L Lei et al. applied a recurrent neural network and autoencoder to encode a patient's hospital records into a low-dimensional dense vector~\cite{lei2018effective}.
Considering the varying length of patients' EHRs, the non-negative tensor factorization models the input sequence to a temporal tensor and serves as the input to the LSTM~\cite{yin2019learning}.
In addition to RNNs, convolutional neural networks (CNNs) are also used for patient representation learning tasks.
Compared with RNN, CNN cannot handle sequence data of indefinite length, so the sequence needs to be preprocessed before being input to the CNN.
In~\cite{suo2018deep}, the patient visit sequence data is used as input to the CNN after being padded to a fixed length,
while in~\cite{xu2018raim}, CNN is used to extract features from the dense ECG waveform and as part of the input to the RNN model.
The above deep learning models can be followed by task-specific output layers for various clinical tasks,
such as mortality prediction~\cite{rajkomar2018scalable}, length-of-stay prediction~\cite{zhou2017learning}, patient subtyping~\cite{baytas2017patient}, medical cost~\cite{stojanovic2016modeling} etc.

\textbf{Perception of uneven time intervals}
The uneven time interval between two adjacent records in the time-series sequence is a typical characteristic of clinical multivariate sequences.
The uneven time interval results from differences in data modality (e.g., waveform data, the medical prescription text, manual record, and medical examination result values) and their observation frequency.
However, Basic RNN cannot be used to perceive uneven time intervals between input records
since the time interval between records in the sequence and the changing rates of the variables are assumed to be consistent by default.
Based on the assumption that the longer the time interval, the lower the reliability of the memory inherited by the RNN from the previous moment,
T-LSTM~\cite{baytas2017patient} uses the time interval between adjacent records after a nonlinear mapping (monotonically decreasing function)
as a discounting factor for the short-term memory of the previous timestep accepted by the LSTM unit.
In DeepCare~\cite{pham2016deepcare}, time parameterization is introduced to handle irregular timing by modifying the forget gate.
Signe Moe et al. ~\cite{moe2021decoupling} shows another route to improve the forward propagation between adjacent units of the RNN along 
the neural ODE proposed in~\cite{chen2018neural} so that it possesses the ability to perceive uneven time intervals between records in a sequence.

The changing rate of variables in real-world data, especially in clinical scenarios, 
is also an important consideration for making correct decisions.
The above studies have attempted to improve the basic RNN and thus deal with the time interval in the sequence, 
but have not yet aimed at directly sensing the rate of change in physiological conditions.

\section{DATA DESCRIPTION}
\label{sec:data-description}
Medical time-series data from EHR contain multiple variables to describe the physiological status of the patients with different dimensions,
and more abundance in types and modalities of clinical variables in the intensive care unit (ICU).
The way these variables are obtained is highly variable,
i.e., the sampling frequencies of waveform data (e.g, electrocardiogram) are higher than 30 Hz
, blood gases (e.g., oxygen saturation, partial pressure of oxygen) are often performed every 10 minutes,
, vital signs (heart rate, pH, and respiratory rate) are performed every 5 minutes, respectively, 
and demographic information (e.g., height, gender, and age) are recorded only once during an admission,
Furthermore, manual recording often causes random missing variables.

\begin{figure}
    \centering
    \includegraphics[width=\linewidth]{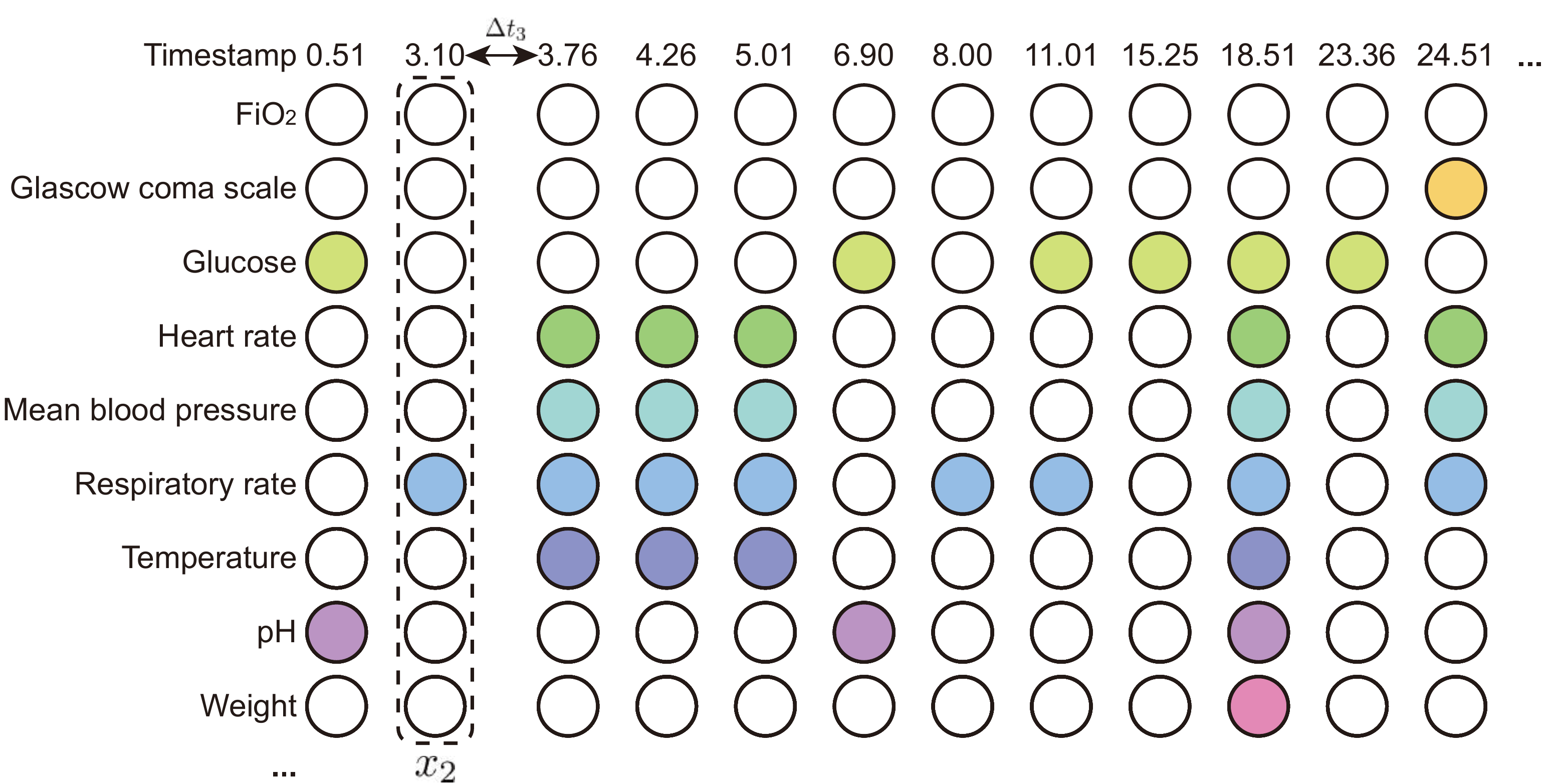}
    \caption{A sample of clinical multivariate time series data extracted from MIMIC-III. 
            Without loss of generality, the variables and timestamps in the sequence are sampled.
            The colored circles indicate that the corresponding variable has a real value collected at that timestamp, 
            while the gray circles indicate that the variable is missing.
            The timestamp is a relative time generated from the time the patient was admitted to the ICU as the starting time point.}
    \label{fig:data-description}
\end{figure}

\begin{figure}
    \centering
    \includegraphics[width=\linewidth]{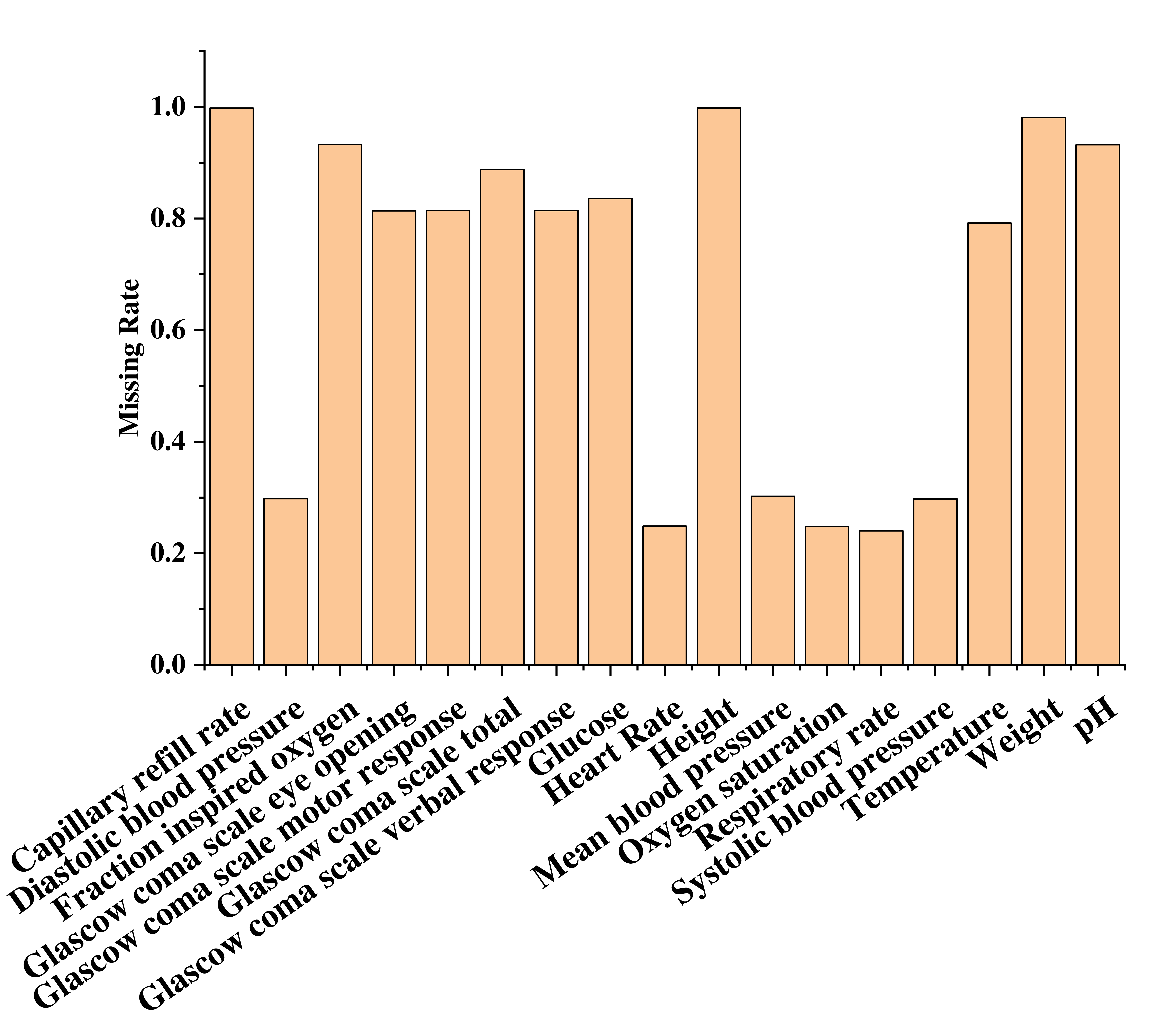}
    \caption{The missing rates of the clinical variables.}
    \label{fig:missing-rate}
\end{figure}

As shown in Figure~\ref{fig:data-description} for an ICU stay sample extracted from MIMIC-III by pipeline provided in \cite{harutyunyan2019multitask},
the clinical variables in the sequence are missing to varying degrees, and even some clinical variables (e.g., $\text{FiCO}_2$) are completely missing.
The mean and standard deviation of the missing rate for each clinical variable obtained from the statistics of all ICU stays are shown in Figure~\ref{fig:missing-rate},
in which the missing rate of these clinical variables is usually higher than $0.5$, or even higher than $0.9$.

Another significant characteristic is that the elapsed time between two records in the sequence is not uniform.
The mean and standard deviation of the time interval between adjacent records at different sampling rates for 
the PhysioNet2012 and MIMIC-III datasets used in this study are shown in Figure~\ref{fig:elasped-time}.
This inhomogeneity is also caused by differences in the timing and frequency of individual 
variable acquisitions, so aligning the values of the variables according to time manifests 
itself as an uneven time interval between records.


\begin{center}
\begin{figure*}
\begin{center}
  \begin{minipage}{1.\linewidth}
    \makebox[.33\linewidth]{\includegraphics[width=.3\linewidth]{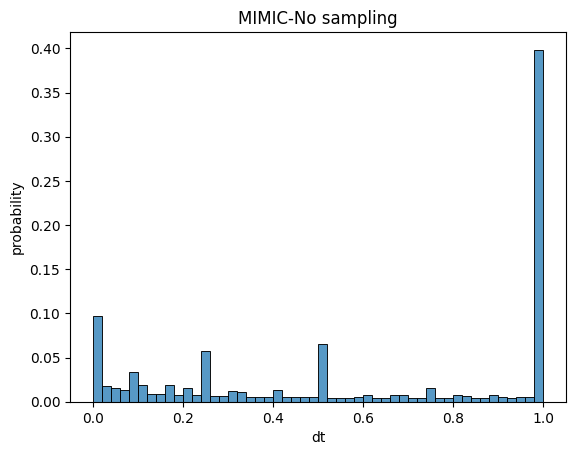}}%
    \makebox[.33\linewidth]{\includegraphics[width=.3\linewidth]{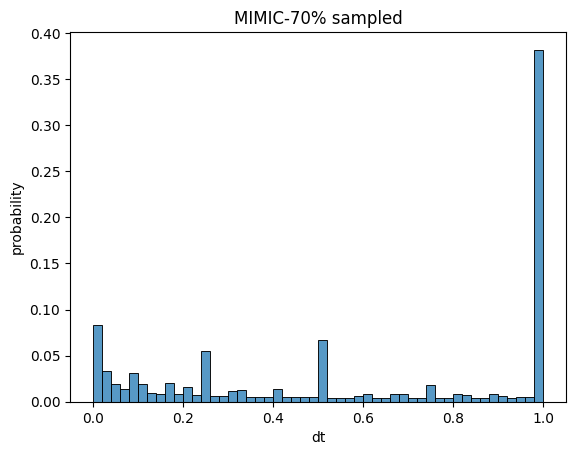}}
    \makebox[.33\linewidth]{\includegraphics[width=.3\linewidth]{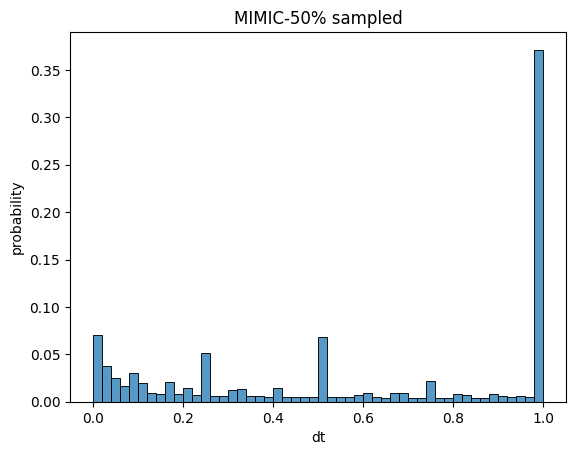}}
    
    \makebox[1.\linewidth]{\small \textbf{MIMIC}}
    
    \medskip

    \makebox[.33\linewidth]{\includegraphics[width=.3\linewidth]{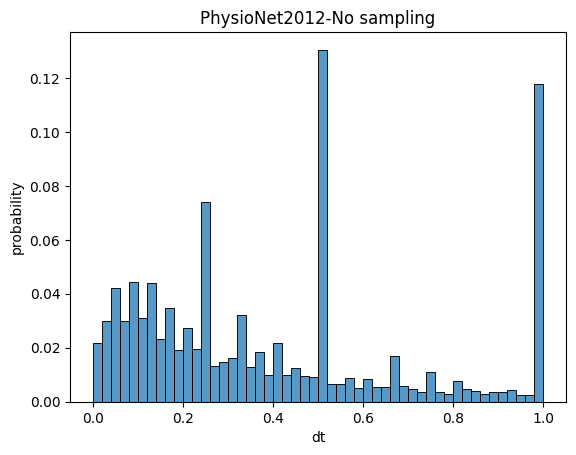}}%
    \makebox[.33\linewidth]{\includegraphics[width=.3\linewidth]{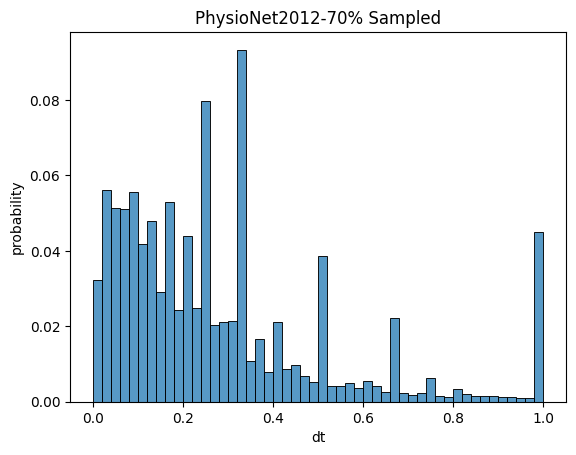}}
    \makebox[.33\linewidth]{\includegraphics[width=.3\linewidth]{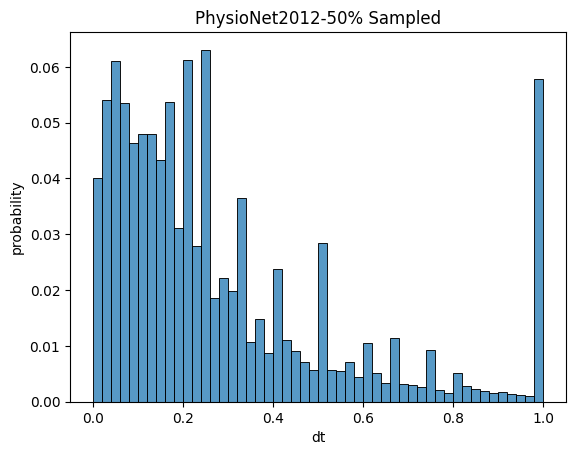}}
    
    \makebox[1.\linewidth]{\small \textbf{PhysioNet2012}}
  \end{minipage}%
  \end{center}
  \caption{Time Interval Distribution}
\end{figure*}
\end{center}

In this study, we describe a clinical time series $S$ with a length of $n$ as a set containing  $n$ records $x$, 
i.e., ${S}=\{x_1 ,{x_2}, \cdots ,{x_n}\}$, $|S|=n$.
Given $D_r$ is the dimension of time series data,
record $x_i \in \mathbb{R}^{D_r}$  can be described as the set of clinical variables that are observed at the time $t_i$.
The elapsed time between two adjacent records is defined as $\Delta t_i=t_i-t_{i-1}, \Delta t_1=1$
which is not assumed consistent in this study. 
The record $x_{t_i}$ is also not assumed to contain complete clinical variables, i.e. $|x_{t_i}| \leq D_r$.

\section{METHOD}
\label{sec:Method}
\begin{figure*}
    \centering
    \includegraphics[scale=0.7]{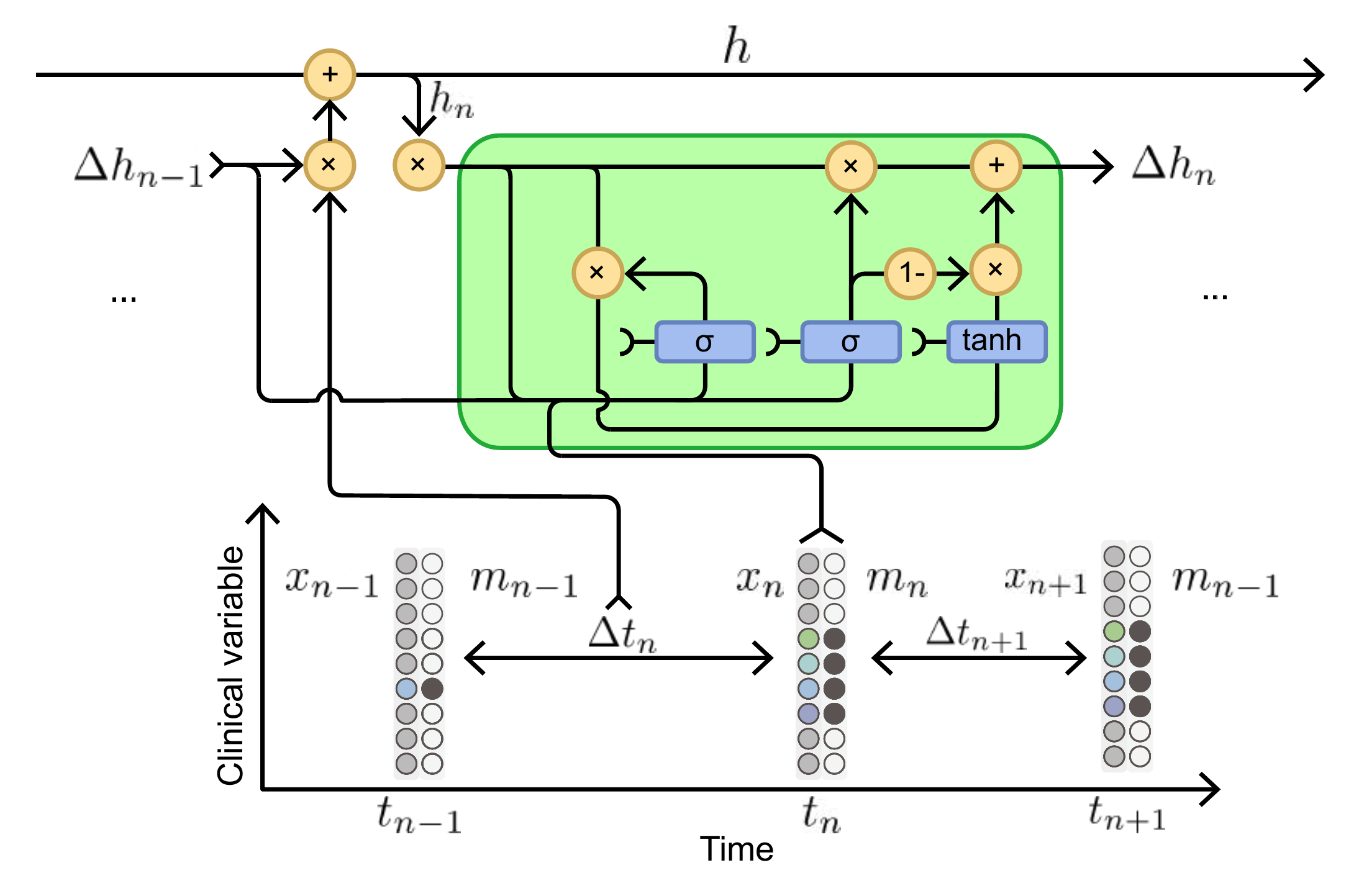}
    \caption{Illustration of GRU-TV Unit}
    \label{fig:model-struct}
\end{figure*}

The overall structure of GRU-TV is shown in Figure~\ref{fig:model-struct}.
It contains two main components based on GRU, which will be described below:
the time interval perception to perceive the time interval between adjacent records in the sequence, 
and the velocity perception to perceive the velocity at which the physiological status representation of the patient corresponding to the input sequence changes.
Our approach is not limited to GRU but can be applied to any RNN model and its variants as a plug-and-play plugin. 

\subsection{Missing value preprocessing}
As mentioned in Section~\ref{sec:data-description}, there are a large number of missing value in clinical multivariate time-series data, 
while GRU takes fixed-length vectors as input, so data padding is widespread in this data.
The missing variables in the series are filled with the latest value, 
that is, the actual value of the filled variable in the previous collection is used as the filling value 
(if the record is missing at the beginning of the series, the default value of the variable is used as the real value).
However, a large number of filling values introduce noise into the input data and change the inherent distribution of the variable,
so the filling values and real values should be distinguished so that the GRU can perceive the difference.

In GRU-TV, we expand the input of the basic standard GRU to include the variable mask vector corresponding to the input record as well.
The forward propagation of the standard GRU is modified as follows:
\begin{equation}
    \begin{gathered}
        r_i = \sigma\left(W_r\left[x_i, h_{i-1}, m_i\right] + b_r\right)\\
        z_i = \sigma\left(W_z\left[x_i, h_{i-1}, m_i\right] + b_z\right)\\
        g_i = tanh\left(W_g\left[x_i, r_t\odot h_{i-1}, m_i\right]\right)\\
        h_i = z_i\odot h_{i-1} + \left(1-z_i\right)\odot g_i
    \end{gathered}
    \label{eq:GRU}
\end{equation}
where $W_r$, $W_z$, and $W_g \in \mathbb{R}^{(D_r \times 2 + D_h) \times D_h}$ are the weights of GRU,
$b_r$, $b_z$ and $b_g \in \mathbb{R}^{D_h}$ are the bias of GRU,
$\sigma\left( \cdot \right)$ is sigmoid activate function, 
and $\odot$ is element-wise multiplication.
\subsection{Time interval perception}
\label{sec:time-interval-perception}
For a sequence with length $n$, RNNs such as GRU update the hidden state by combining the transformation of the previous hidden state and the current input record:
\begin{equation}
    h_{i+1}=\text{GRU}(h_i,x_i,\theta_i)
\end{equation}
where $i \in \left\{0,\dots, n\right\}$ and $h_i \in \mathbb{R}^{D_h} $. 
This transformation is discrete and does not take into account the time interval between adjacent records, 
and therefore has natural drawbacks for patient representation learning using multivariate clinical data.
To alleviate this deficiency, we would like GRU to update the hidden state as follows:
\begin{equation}
    h_{i+1}=\text{GRU}^{'}(h_i,x_i,t_i, t_{i-1}, \theta_i)
\end{equation}
where $t_i$ and $t_{i-1}$ are the timestamps of records $x_i$ and $x_{i-1}$, respectively.

Inspired by the neural ODEs proposed in~\cite{NEURIPS2018_69386f6b},
instead of expanding an additional elapsed time as the input to the GRU cell, 
we keep the forward propagation of the standard GRU cell unchanged in this step
but improve the update method of the hidden state $h_t$ after the cell receives a record.
In the proposed GRU-TV, 
the hidden representation $h_n$ is not fed into GRU directly in the next timestep, 
but we obtain the differential equation for $h_t$:
\begin{equation}
    \begin{aligned}
        \Delta h_i & = h_i - h_{i-1}\\
        & =\left(1-z_i\right) \cdot \left(g_i - h_{i-1}\right)
    \end{aligned}
    \label{eq:delta-h}
\end{equation}
which leads to the update of $h_i$
\begin{equation}
    h_i = h_{i-1} + \Delta t_i \cdot \Delta h_i
    \label{eq:update-h}
\end{equation}
where $\Delta t_i = t_i - t_{i-1}$ is the elapsed time of $x_i$ and $x_{i-1}$.

\subsection{Velocity perception}
In clinical practice, the instantaneous rate of vital signs associated with physiological kinetics is an important indicator of a patient's physiological status. 
How to parameterize the instantaneous rate of a patient's physical condition by GRU?
In the proposed GRU-TV, the ODE of the hidden state is also used as an input thus giving 
the GRU the ability to perceive the velocity.
As proved in~\cite{NEURIPS2018_69386f6b,de2019gru}, when the time interval between adjacent records is infinitesimal, the limiting form of the hidden state in Section~\ref{sec:time-interval-perception}, 
i.e., its ODE is:
\begin{equation}
    \frac{dh\left(t\right)}{dt}=\left(1-z\left(t\right)\right) \cdot \left(g\left(t\right)-h\left(t\right)\right)
\end{equation}
The ODE of $h_t$ stays within the $\left[-1, 1\right]$ range, which guarantees the convergence of our model.

Although the instantaneous rate of $h$ exists, we still use its discrete form as an input to GRU in this study for computational convenience:
\begin{equation}
    \begin{gathered}
        r_i = \sigma\left(W_r\left[x_i, h_{i-1}, m_i, \Delta h_i\right] + b_r\right)\\
        z_i = \sigma\left(W_z\left[x_i, h_{i-1}, m_i, \Delta h_i\right] + b_z\right)\\
        g_i = tanh\left(W_g\left[x_i, r_i\odot h_{i-1}, m_i, \Delta h_i\right] + b_g\right)\\
        \Delta h_i  = \left(1-z_i\right) \cdot \left(g_i - h_{i-1}\right) \\
        h_i = h_{i-1} + (t_i - t_{i-1}) \cdot \Delta h_i
    \end{gathered}
    \label{eq:GRU-DH}
\end{equation}
where $\Delta h_i \in \mathbb{R}^{D_h}$ is the discrete form of $dh$, which is defined earlier, $W_r$, $W_z$, and $W_g \in \mathbb{R}^{(D_r\times 2 +D_h \times 2)\times D_h}$ are the weight.

It is worth noting that here we use the instantaneous rate of the hidden state rather than the input variable as the velocity for the perception module. 
The reason is that the velocity of the input variable cannot be accurately obtained because there are a large number of filled values in input sequence, 
while the hidden state can better represent the patient's physiological condition.

\subsection{Loss funcion}
The hidden attitudes obtained after traversing the entire input sequence are fed 
into a task-specific separator, which in this study is a fully-connected neural network
with a sigmoid function as the activation function.
Binary cross entropy loss (BCE) loss was selected as the loss function 
since the experiments in this study are multi-label binary classification tasks. 
The BCE loss is defined as:
\begin{equation}
    \mathcal{L} =\frac{1}{K}\sum^{K}_{k=1}y_k\cdot \log\left(\hat{y}_k\right)+\left(1-y_k\right)\cdot \log\left(1-\hat{y}_k\right)
\end{equation}
where $y_k \in \left\{0, 1\right\}$ is the label of $k^{th}$ binary classification task, 
and $\hat{y} \in \left(0, 1\right)$ is the predicted probability of $k\text{-th}$ classification task.

\subsection{Sequence sampling}
The scale of sequential sampling is record, i.e., for each time series in the dataset, 
a certain percentage of records are selected to construct the post-sampling time series.
In this study, two sequence sampling methods are applied to verify the performance of the algorithm on data sets with uneven time intervals.
Native random sampling was applied to the Physionet2012 dataset. Any record in the sequence was selected with equal probability because of the uniform distribution of time intervals between the adjacent records.
As shown in the figure, the most time intervals between adjacent records are 1 hour in MIMICIII dataset. An inverse proportional sampling method shown in Algorithm~\ref{alg:1} was applied so that the time interval between adjacent records can be evenly distributed.

In Algorithm~\ref{alg:1}, the sampling scale dictionary $P$ is generated as follows:
\begin{equation}
    \begin{aligned}
         p_t &= \frac{1}{c_t \times C_{all}}\\
         C_{all} &= \sum_{t \in \omega}c_t
    \end{aligned}  
\end{equation}
where $p_t$ is the probability that the record with time interval $t$ from the previously selected record is selected,
$c_t$ is the number of time intervals $t$ between adjacent records in the original dataset.
Fewer time intervals in the original dataset are more likely to be sampled.
Sampling stops when the rate of sampled records to the records in the original data set reaches the target rate.

\begin{algorithm}
    \label{alg:1} 
    \DontPrintSemicolon
    \SetAlgoLined
    \SetKwInOut{Input}{input}\SetKwInOut{Output}{output}
    \Input{
        Sampling probability dictionary: $P={<t, p_t>}$,\\ ~Sequence dataset: $D={S}$,\\ ~Target sample rate: $T$ \;
    }
    \Output{
        Selected index dictionary: $I={<S, {idx}>}$ \;
    }
    \For{$S \in D$}{
        $D[S] \leftarrow [0]$\;
    }
    $R\leftarrow0$\;
    \While{$R < T$}{
        \For{$S \in D$}{
            \For{$i \leftarrow 1 \KwTo LEN(S)$}{
                \If{$i \in I[S]$}{
                    CONTINUE\;
                }
                $t \leftarrow TimeInterval(S[i], I[S])$\;
                $p \leftarrow P[t]$\;
                $r \leftarrow RandomNum(0, 1)$\;
                \If{$r \leq p$}{
                    $AppendIndex(I[S], i)$\;
                }
            }
        }
        $R \leftarrow CalSampleRate(D, I)$
    }
    \caption{Inverse proportional sequence sampling}
\end{algorithm}


\section{EXPIREMENT}
\label{sec:EXPIREMENT}
\subsection{Comparing model}
We compare our proposed method to the following deep learning methods: GRU-Standard, GRU-Decay~\cite{che2018recurrent}, LSTM~\cite{hochreiter1997long}, 
and T-LSTM~\cite{baytas2017patient}.
We also take the only-time-aware included model (GRT-T) as a comparison model
and the $\gamma_h$ and $\gamma_x$ used in GRU-Decay are also embedded in GRU-T respectively for more comprehensive comparisons.
For $\text{GRU-T-}\gamma_h$, the $h_i$ calculated in Equation~\ref{eq:GRU} is modified by $\gamma_h$:
\begin{equation}
    \begin{aligned}
        \gamma_{h_i} &= e^{\left\{-max\left(0, W_{\gamma_h}\delta_t+b_{\gamma_h}\right)\right\}}\\
        h_i &\leftarrow h_i \gamma_{h_i}
    \end{aligned}
    \label{eq:ht}
\end{equation}
where $\delta_t$ is a vector containing the elapsed time of the last collection for every variable.
For $\text{GRU-T-}\gamma_x$, the trainable decay coefficient is applied in the filling of the missing values in record $x_t$:
\begin{equation}
    \begin{aligned}
        \gamma_{x_i} &= e^{\left\{-max\left(0, W_{\gamma_x}\delta_i+b_{\gamma_x}\right)\right\}}\\
        \hat{x}^d_t &= m^d_i x^d_i + \left(1-m^d_i\right)(\gamma^d_{x_i} x^d_{last}+\left(1-\gamma^d_{x_i}\right)\widetilde{x}^d)\\
    \end{aligned}
\end{equation}
where $\widetilde{x}^d$ is the default value of ${\text{d-th}}$ (the average of real value is used as default in this study).
For $\text{GRU-T-}\gamma_{h+x}$, both the $\gamma_h$ and $\gamma_x$ in GRU-Decay are applied in GRU-T.
\subsection{Dataset}
We demonstrate the performance of our proposed models on two real-world EHR datasets: 
PhysioNet2012\footnote{https://www.physionet.org/content/challenge-2012/1.0.0/} and MIMIC-III\footnote{https://mimic.mit.edu/},
and both of them are publicly available.

\textbf{PhysioNet2012} is a dataset for the PhysioNet Computing in Cardiology Challenge 2012. 
This multivariate clinical time series dataset consists of records from 12,000 ICU stays, 
all patients were adults who were admitted for a wide variety of reasons to 
cardiac, medical, surgical, and trauma ICUs. ICU stays of less than 48 hours have been excluded from this study.

\textbf{MIMIC-III}  is a large, freely-available dataset comprising health-related data associated with over 40,000 patients who stayed in critical care units of the Beth Israel Deaconess Medical Center between 2001 and 2012.
Over 58,000 hospital admission records were collected in this dataset.

For each dataset, after organizing the sporadically recorded data points into multivariate time series $S$,
the variable time interval sequence $\Delta$ and mask sequence $M$ corresponding to $S$ was obtained by referring to the pipeline in GRU-Decay~\cite{che2018recurrent}.
In addition, the elapsed time between adjacent records $x_{t_i}$ and $x_{t_{i-1}}$ in
the elapsed time sequence required for GRU-T was obtained by the difference between the corresponding time points $t_i$ and $t_{i-1}$, 
where the elapsed time between $x_1$ and the previous record was set to $1$.

\subsection{Experimental Setup}
To permit a fair comparison between the methods,
consistent hyperparameters, network setting, training pipeline, and dataset splits were used for all the deep learning methods.
The training was stopped if the training epoch is greater than $30$ and there is no performance improvement 
on the validation set for $3$ consecutive iterations.
The parameters with the best validation performance were selected to evaluate the performance on the test dataset.
The final performance was the average of three independent test performances.
The comparison metric for this study was the area under the ROC curve (AUROC).
The macro average AUC was also calculated for the multi-task classification. 

The hidden states, obtained after traversing all records in the input sequence, 
are fed to a fully connected layer to obtain predictions. 
The sigmoid activation function was used for multi-label binary classification.

In this study, the current optimal models GRU, GRU-Decay and T-LSTM, which are widely used for patient representation learning 
using clinical sequences, are used as comparison models.
Furthermore, ablation experiments are also conducted to evaluate the performance improvement of each module.

\subsection{Multi-task classification on PhysioNet2012}
In this study, PhysioNet2012 dataset was used for multi-task classification, 
which contains four sub-tasks: in-hospital mortality prediction, 
length-of-stay classification, 
cardiac condition classification, 
and surgery recovery prediction.

In this task, the non-sequential variables in the dataset were discarded due to the basic GRU unit requires the input of sequence data.
The variables of the time series were reorganized,
in which variables collected at the same point in time or those collected at intervals of less than 5 minutes were aligned in the same record. 


\begin{table*}[htbp]
    \centering
    \caption{Performances comparison of methods on multi-task classification of complete PhysioNet2012 dataset, and sampled PhysioNet2012 dataset with 70\% and 50\% sampling rate. 
        AUROC: area under the receiver operating characteristic curve,
        AUPRC: area under the precision\-recall curve,
        Mor: in-hospital mortality,
        Los: length-of-stay less than 3 days,
        Car: whether the patient had a cardiac condition,
        Sur: whether the patient was recovering from surgery,
        Mac: macro average AUCROC.
        The best and second-best performance for each subtask as well as the average are bolded and underlined.
        }
      \begin{tabular}{clp{12mm}p{12mm}p{12mm}p{12mm}p{12mm}}
      \toprule[1.5pt]
      \multirow{2}{*}{Sampling Rate}& \multirow{2}{*}{Method} & \multicolumn{5}{c}{AUROC} \\
      \cmidrule(lr){3-7}
      \multirow{5}{*}{100\%} & & \multicolumn{1}{l}{Mor} & \multicolumn{1}{l}{LoS} & \multicolumn{1}{l}{Car} & \multicolumn{1}{l}{Sur} & \multicolumn{1}{l}{Mac} \\
      \midrule[0.5pt]
      &GRU-Standard     &0.8638 &0.8043&\underline{0.9510}&0.8956&0.8787\\
      &GRU-Decay  &0.8601&0.8199&\textbf{0.9525}&\textbf{0.9043}&0.8842\\
      &T-LSTM       &0.8469&0.7639&0.9274&0.8735&0.8529\\
      \cmidrule(lr){2-7}
      &$\text{GRU-T-}\gamma_h$&0.8641&\textbf{0.8432}&0.9507&0.8992&\textbf{0.8893}\\
      &$\text{GRU-T-}\gamma_x$&\underline{0.8667}&0.8200&0.9497&0.8947&0.8827\\
      &$\text{GRU-T-}\gamma_{h+x}$&\textbf{0.8864}&0.8086&0.9438&0.8980&0.8842\\
      &GRU-T  &0.8619&0.8255&0.9499&0.9019&\underline{0.8848}\\
      &GRU-TV  &0.8498&\underline{0.8380}&0.9475&\underline{0.9028}&0.8845\\
      \midrule[0.5pt]
      \multirow{5}{*}{70\%} &GRU-Standard&0.6709&0.7254&0.9170&0.8046&0.7795\\
      &GRU-Decay  &0.6488&0.7774&0.8492&0.7461&0.7554\\
      &T-LSTM       &\underline{0.7512}&\underline{0.7924}&0.9097&\underline{0.8215}&\underline{0.8187}\\
      \cmidrule(lr){2-7}
      &$\text{GRU-T-}\gamma_h$&0.7371&0.7808&0.8987&0.8074&0.8060\\
      &$\text{GRU-T-}\gamma_x$&0.7085&0.7699&0.8914&0.8063&0.7940\\
      &$\text{GRU-T-}\gamma_{h+x}$&0.7449&0.7716&0.9026&0.8005&0.8049\\
      &GRU-T  &0.7505&0.7698&\underline{0.9194}&0.8135&0.8133\\
      &GRU-TV  &\textbf{0.7891}&\textbf{0.8017}&\textbf{0.9208}&\textbf{0.8539}&\textbf{0.8413}\\
      \midrule[0.5pt]
      \multirow{5}{*}{50\%} &GRU-Standard     &0.7132&0.7160&0.8957&0.8102&0.7838\\
      &GRU-Decay  &0.7426&\textbf{0.7857}&0.8819&0.8093&0.8049\\
      &T-LSTM       &0.7666&0.7082&\underline{0.9133}&\underline{0.8291}&0.8043\\
      \cmidrule(lr){2-7}
      &$\text{GRU-T-}\gamma_h$&0.7425&0.7357&0.8942&0.8062&0.7946\\
      &$\text{GRU-T-}\gamma_x$&0.7222&0.7395&0.9013&0.8026&0.7914\\
      &$\text{GRU-T-}\gamma_{h+x}$&0.7248&\underline{0.7724}&0.9032&0.8122&0.8031\\
      &GRU-T  &0.7453&0.7712&0.9045&0.8183&\underline{0.8098}\\
      &GRU-TV  &\textbf{0.7792}&0.7683&\textbf{0.9193}&\textbf{0.8467}&\textbf{0.8283}\\
      \bottomrule[1.5pt]
      \end{tabular}%
    \label{tab:physionet2012}%
  \end{table*}%
Table~\ref{tab:physionet2012} depicts the experimental results on the multi-task classification of PhysioNet2012 dataset.
For each metric, the best and second best are marked in bold and underlined, respectively.

In general, the application of ODE improves the performance of the standard GRU for each subtask,
and this improvement is more obvious in the sampled simulation data.
The velocity perception was more significant for performance improvement compared to time interval perception, especially on sampled datasets.

When the full sequences are used,
The GRU-T model combined with $\gamma_h$ term or $\gamma_x$ term performs best on in-hospital mortality and length-of-stay less than 3 days prediction tasks,
while on cardiac and surgery classification tasks, GRU-Decay performs better. 
In terms of average performance, GRU-T combined with $\gamma_h$ term and GRU-T are optimal and sub-optimal models, respectively.
When simply comparing GRU-T and GRU-Decay, 
the former has a more pronounced improvement relative to the GRU-Standard model. 
The velocity perception module of GRU-TV does not improve the performance of each subtask compared to GRU-T.

In the sampled datasets, 
perception of the time interval between records is more important since random sampling results in a more uneven time interval between records in a sequence.
The importance of the ability to perceive the unevenness of the time interval between records can be demonstrated by the improved performance of T-LSTM and GRU-T on each subtask relative to the other models.
Except for the length of stay classification subtask, 
the $\gamma_h$ and $\gamma_x$ terms in GRU-Decay can even have a negative impact on the performance whether they are applied to GRU-Standard or GRU-T.

When evaluating using the sampled datasets,
GRU-TV significantly outperforms the other models in every subtask.
Compared to the suboptimal models, the AUC of GRU-TV improves by 2.8\% and 2.3\% on 70\% and 50\% sampled datasets, respectively.

The average macro AUC of GRU-T and GRU-TV in the three data sets are $0.8360$ and $0.8514$, which are both better than the $0.8253$ of T-LSTM.
This shows that our proposed GRU-T and GRU-TV are generalizable and robust on both sampled and complete datasets.

\subsection{Acute phenotype classification on MIMIC-III}
In the ICU setting, acute illnesses with short onset cycles and diagnostic time windows are more difficult to
diagnose clinically and require more rapid intervention and treatment.
In this study, the proposed model is evaluated by the acute phenotype subtyping task using the MIMIC dataset.
Acute and unspecified renal failure, acute cerebrovascular disease, pulmonary collapse, pneumonia,
respiratory failure, septicemia, and shock are included in the acute phenotypes.

When the complete dataset is used, GRU-Decay always performs best both on individual subtasks and overall performance.
However, GRU-Decay's $\gamma_h$ and $\gamma_x$ terms applied to GRU-T can not improve the performance but have a negative effect.

On the 20\% sampled dataset, the reduction in performance of the models is not significant, 
despite the fact that only 20\% of the records are used for patient representation learning.
Among them, the macro AUC of the GRU-Decay model decreased from 0.8137 to 0.7664 (5.81\% reduction), while the macro AUC of the GRU-TV model decreased from 0.8050 to 0.7800 (3.11\% reduction).GRU-TV is also significantly more robust than GRU-Decay on each subtask.

\begin{table*}[htbp]
    \centering
    \caption{Performances comparison of acute phenotype classification using complete sequences, and sampled sequences with 70\% and 50\% sampling rate extracted from  MIMIC-III dataset.
            AURF: acute and unspecified renal failure,
            ACD: acute cerebrovascular disease,
            PC: pulmonary collapse,
            PN: pneumonia
            RF: respiratory failure,
            SE: septicemia,
            SH: shock.
            The best and second-best performance for each subtask as well as the average are bolded and underlined.
    }
      \begin{tabular}{clp{12mm}p{12mm}p{12mm}p{12mm}p{12mm}p{12mm}p{12mm}p{12mm}}
        \toprule[1.5pt]
        \multirow{2}[2]{*}{\textbf{Sampling Rate}}&\multirow{2}[2]{*}{\textbf{Method}} & \multicolumn{8}{c}{\textbf{AUPRC}} \\
      \cmidrule(lr){3-10}
            &   & \textbf{AURF} & \textbf{ACD} & \textbf{PC} & \textbf{PN} & \textbf{RF} & \textbf{SE} & \textbf{SH} & \textbf{MACRO} \\
      \midrule
      \multirow{5}[2]{*}{100\%} & GRU-Standard&\underline{0.7619}&\underline{0.8830}&\underline{0.6746}&0.7835&0.8710&0.8070&\textbf{0.8621}&\underline{0.8061}\\
        & GRU-Decay &\textbf{0.7630}&\textbf{0.9050}&\textbf{0.6901}&\textbf{0.7909}&\textbf{0.8794}&\textbf{0.8098}&0.8577&\textbf{0.8137}\\
        & T-LSTM  &0.7148&0.7945&0.6545&0.7695&0.8372&0.7931&0.8326&0.7709\\
        \cmidrule(lr){2-10}
        & $\text{GRU-T-}\gamma_h$    &0.7503&0.8544&0.6709&0.7760&0.8685&0.7930&0.8533&0.7952\\
        & $\text{GRU-T-}\gamma_x$    &0.7233&0.8317&0.6530&0.7656&0.8517&0.7821&0.8340&0.7774\\
        & $\text{GRU-T-}\gamma_{h+x}$    &       &       &       &       &       &       &       &  \\
        & GRU-T    & 0.7567 &0.8717&0.6726&\underline{0.7892}&0.8733&0.8052&0.8582&0.8039\\
        & GRU-TV    &0.7577&0.8817&0.6704&0.7835&\underline{0.8747}&\underline{0.8073}&\underline{0.8598}&0.8050\\

      \midrule
      \multirow{5}[2]{*}{20\% (new method)} & GRU-Standard&  &   &   &   &   &   &   &\\
        & GRU-Decay &0.7162&0.8040&0.6530&0.7637&0.8394&0.772&0.8169&0.7664\\
        & T-LSTM&  &   &   &   &   &   &   &\\
        \cmidrule(lr){2-10}
        & $\text{GRU-T-}\gamma_h$&  &   &   &   &   &   &   &\\
        & $\text{GRU-T-}\gamma_x$&  &   &   &   &   &   &   &\\
        & $\text{GRU-T-}\gamma_{h+x}$&  &   &   &   &   &   &   &\\
        & GRU-T&  &   &   &   &   &   &   &\\
        & GRU-TV&0.7278&0.8469&0.6624&0.7689&0.8458&0.7747&0.8334&0.7800\\
    \bottomrule[1.5pt]
    \end{tabular}%
    \label{tab:addlabel}%
  \end{table*}%

\section{PATIENT REPRESENTATION} 
\begin{center}
\begin{figure*}
\begin{center}
  \begin{minipage}{1.\linewidth}
    \makebox[.33\linewidth]{\includegraphics[width=.33\linewidth]{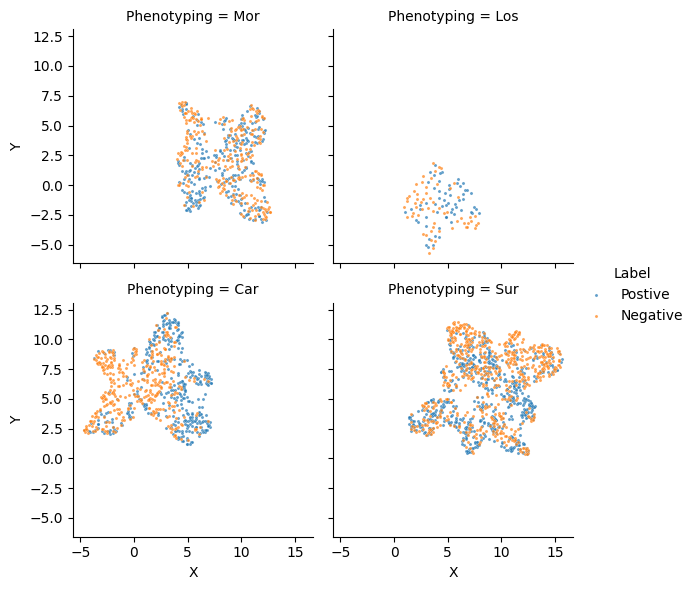}}%
    \makebox[.33\linewidth]{\includegraphics[width=.33\linewidth]{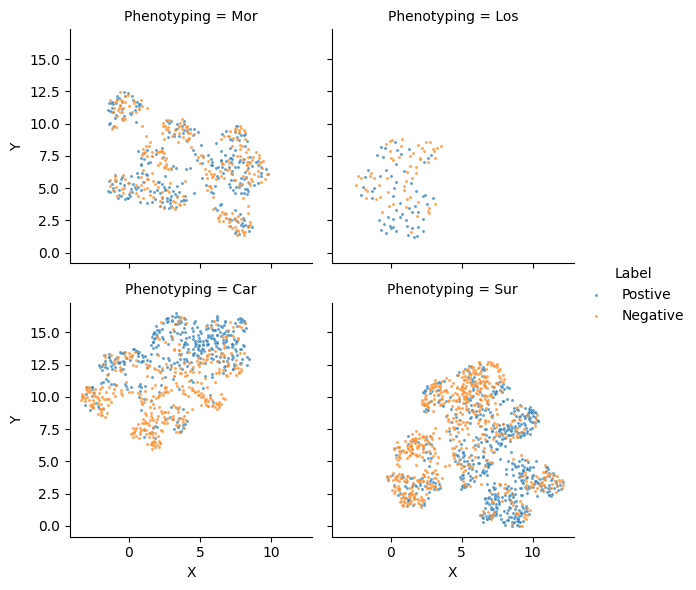}}
    \makebox[.33\linewidth]{\includegraphics[width=.33\linewidth]{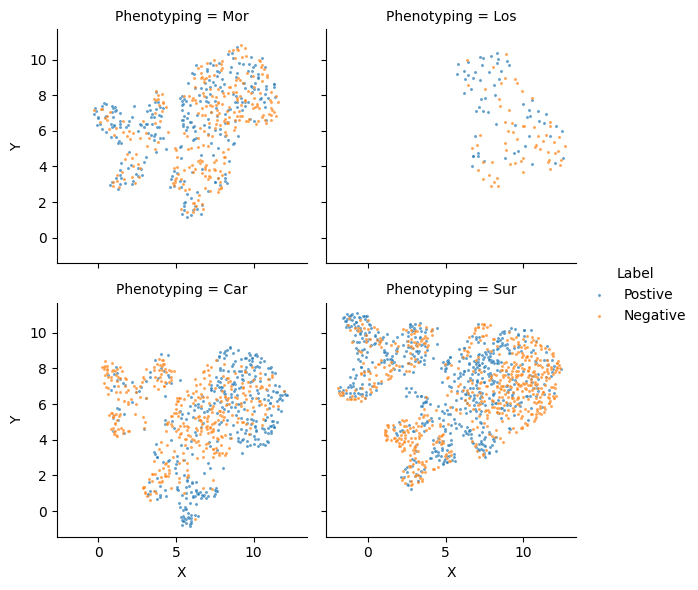}}
  \end{minipage}%
  \end{center}
  \caption{The UMAP Visualizations of Patient Representation}
  \label{fig:patient-representation}
\end{figure*}
\end{center}

As shown in Figure~\ref{fig:patient-representation}, TODO

\section{CONCLUTION}
\label{sec:CONCLUTION}
In this study, we proposed a GRU variant unit for modeling clinical multivariate time-series data, called GRU-TV. 
GRU-TV improves the forward propagation between adjacent GRUs through the ODE and velocity of the hidden state to 
represent the changing process of the patient's physiological status in a continuous manner.

Compared to RNN models such as GRU and LSTM, GRU-TV allows the time interval between records of the input sequence to 
be of free length, and eliminating the need for imputation.
GRU-TV achieves state-of-the-art results on multiple tasks on both real-world datasets, 
especially on sampled simulated data.

The experimental results show that the proposed GRU-TV model is more advantageous in sequences with uneven time intervals, 
which makes GRU-TV more suitable for applications with low frequency and unstable sampling scenarios.
Furthermore, the  ODE and velocity representation  of the hidden state in GRU-TV can be used as a plug-and-play plugin for other RNN models and achieve performance improvement.
In our future work, it will be applied to other specific tasks of patient representation with
clinical time-series data, to embed it in online real-time alert systems to improve computer-aided diagnostics, 
and to evaluate the generalizability of our model with real-world data.

\section{ACKNOWLEDGMENT}
\label{sec:ACKNOWLEDGMENT}
*
\clearpage
\bibliography{references}

\end{document}

%% file: defs.tex
\SetKwData{OPEN}{\texttt{OPEN}}
\SetKwData{FOCAL}{\texttt{FOCAL}}
\SetKwData{SUBOPEN}{\texttt{SUBOPEN}}
\SetKwData{NONFOCAL}{\texttt{NONFOCAL}}
\SetKwData{CLOSE}{\texttt{CLOSE}}
\newcommand{\JB}[1]{\textbf{\color{blue} JB: #1}}
\newcommand{\PA}[1]{\textbf{\color{red} PA: #1}}
\newcommand{\MG}[1]{\textbf{\color{green} MG: #1}}
\newcommand{\Paragraph}[1]{\smallskip\noindent\textbf{#1}~~}